\documentclass[twocolumn]{article}
\usepackage[english]{babel}
\usepackage{overpic}
\usepackage[letterpaper,top=2cm,bottom=2cm,left=3cm,right=3cm,marginparwidth=1.75cm]{geometry}
\usepackage{url}
\usepackage{booktabs}
\usepackage{multirow}
\usepackage[table]{xcolor}
\usepackage{colortbl}
\usepackage{amsmath}
\usepackage{graphicx}
\usepackage[colorlinks=true, allcolors=blue]{hyperref}

\title{MUXQ: Mixed-to-Uniform Precision MatriX Quantization via Low-Rank Outlier Decomposition}

\author{
\textbf{Seoungsub Lee}$^{*}$, In Seo Kim, and Seon Wook Kim\\
School of Electrical Engineering, Korea University, Seoul, Korea\\
\texttt{\{tmdtjq789@korea.ac.kr, kiminse123@korea.ac.kr, seon@korea.ac.kr}\\
$^{*}$First author
}

\begin{document}
\maketitle

\begin{abstract}
Large language models (LLMs) have achieved outstanding performance across a wide range of natural language processing tasks, but their enormous parameter counts impose substantial memory and computational overheads. This challenge is particularly critical in NPU-based on-device environments, where FP16/FP32 computation is inefficient and integer (INT) quantization is therefore essential. However, existing methods, including ZeroQuant, LLM.int8(), and SmoothQuant, do not fully address input-activation outliers and the associated hardware inefficiencies.

To overcome these limitations, we propose MUXQ (Mixed-to-Uniform Quantization). MUXQ detects outlier channels in input activations and introduces a small auxiliary matrix that redistributes outlier magnitudes across channels, thereby alleviating the outlier problem. This enables even activation outliers to be quantized at low-precision INT levels while preserving a hardware-friendly computation structure. Experiments on GPT-2 models at three scales (0.1B, 0.3B, and 0.7B parameters) using the WikiText-2 dataset show that MUXQ consistently achieves lower perplexity than naive quantization. In particular, under per-tensor quantization, MUXQ quantizes both activations and weights to INT8 while maintaining accuracy close to that of FP16. With only modest computational overhead, MUXQ enables stable low-precision inference and can be readily combined with other quantization techniques. These results suggest that MUXQ provides a promising direction for efficient and accurate LLM inference on edge devices.
\end{abstract}
\vspace{0.5em}
\noindent \href{https://github.com/GillchLee/MUXQ}{https://github.com/GillchLee/MUXQ}

\section{Introduction}
Large language models (LLMs) have recently demonstrated remarkable performance not only in natural language processing, but also in a wide range of applications such as question answering and summarization. As the demand for LLMs continues to grow, users increasingly require models that are both fast and accurate. Transformer-based LLMs typically rely on billions of parameters to achieve high accuracy during training and inference. However, when real-time inference is required, the number of parameters cannot be increased indefinitely for the sake of accuracy alone \cite{ref1}. In addition, the growth in model size has steadily increased the memory usage of hardware accelerators such as graphics processing units (GPUs) \cite{ref2}, making real-time inference even more challenging in on-device environments.

Recently, the use of AI on edge devices, including smartphones and embedded systems, has expanded rapidly. Accordingly, the importance of techniques that can compress LLMs originally executed with floating-point arithmetic (FP16 and FP32) and enable efficient inference using integer (INT) operations has grown significantly \cite{ref3}. This is particularly important in the case of neural processing units (NPUs), which support on-device AI acceleration and are especially optimized for INT computation. In such environments, directly implementing floating-point computation is not only inefficient, but can also impose severe limitations on inference.

\begin{figure*}[t]
    \centering
    \includegraphics[width=0.48\textwidth]{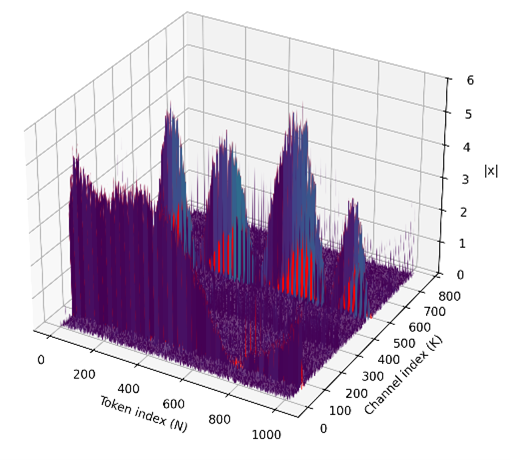}
    \hfill
    \includegraphics[width=0.48\textwidth]{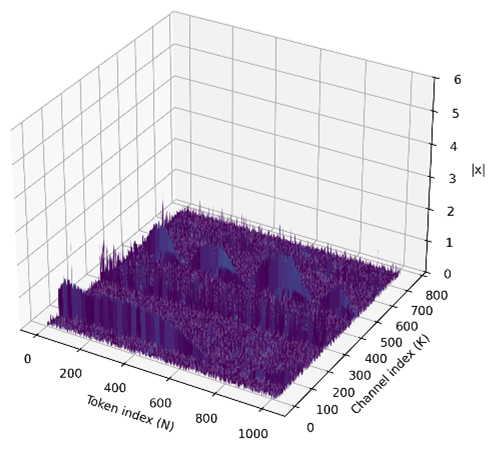}
    \caption{Figure 1. (Left) Activation outliers are concentrated in a small number of channels. (Right) After applying MUXQ, the magnitudes of the outlier channels are reduced.}
    \label{fig:two_figures}
\end{figure*}
Memory bottlenecks have also emerged as one of the major challenges in current LLM architectures. In particular, as model size increases, the KV cache can occupy more than half of GPU memory, which makes practical deployment increasingly difficult \cite{ref2}. Therefore, it has become essential to replace floating-point computation with quantization techniques that compress models into low-precision integer formats, such as INT4 and INT8, in order to alleviate memory bottlenecks and enable efficient inference. Quantization is a technique that reduces memory usage and improves computational efficiency by converting model weights and activations into lower-precision representations \cite{ref4}. For example, when FP16 is quantized to INT8, memory usage can be reduced by half, while general matrix multiplication (GEMM) can theoretically be accelerated by more than 2×. However, outliers in activation matrices make such quantization difficult. As shown in Fig. 1, these outliers are concentrated along only a few specific channels. If they are structurally removed or ignored, model accuracy degrades severely \cite{ref5}.

To address this problem, various quantization methods have been proposed, including LLM.int8() \cite{ref6} and SmoothQuant \cite{ref5}. LLM.int8() prevents significant accuracy degradation by separating outlier features and processing them in FP16 through mixed-precision decomposition. However, the remaining FP16 operations introduce irregular memory access patterns on hardware, which reduce hardware efficiency, and mixed-precision execution itself also lowers computational efficiency. SmoothQuant, on the other hand, reduces the difficulty of quantization by migrating the scaling factors of activations into the weights, but it does not fully resolve the fact that activation outliers still remain much larger than normal elements.

In this way, prior studies have attempted to minimize accuracy loss through mixed-precision computation or scaling-based mitigation. Nevertheless, they either fail to preserve hardware efficiency or leave activation outliers insufficiently addressed. Therefore, in LLM inference, detecting and handling activation outliers is a more critical issue for preserving performance than weight quantization, which is already relatively amenable to low-precision quantization.

In this paper, we propose MUXQ (Mixed-to-Uniform Precision Matrix Quantization) to fundamentally address this issue. MUXQ is a new quantization framework that removes the hardware bottlenecks of conventional mixed-precision schemes (FP16+INT8) and enables uniform low-precision INT8 quantization. Specifically, MUXQ detects activation outliers and adds a small auxiliary matrix to the input activation matrix of each layer. This auxiliary matrix splits and redistributes the magnitudes of the outlier channels, thereby alleviating extreme activation values. As a result, the original activation matrix becomes easier to quantize at low precision, and the overall activation distribution becomes more balanced, allowing high accuracy to be preserved using only INT precision. The right side of Fig. 1 illustrates how MUXQ produces a more balanced activation distribution.

Because all computations in MUXQ are performed in INT precision, it operates in a hardware-friendly manner on accelerators such as NPUs and eliminates the problem of irregular memory access. Although MUXQ introduces a small amount of additional memory usage and computational overhead, it enables stable inference at lower bit precision, such as INT8 or INT4, while reducing both overall system latency and memory usage. In our experiments, MUXQ achieved accuracy comparable to that of mixed-precision LLM.int8() on the 0.1B GPT-2 model, while successfully enabling uniform per-tensor INT8 quantization.

The contributions of this paper are as follows. First, we propose a new low-precision quantization approach that introduces an auxiliary matrix into the activation matrix to mitigate activation outliers. Second, MUXQ can be integrated with other quantization techniques and used in combination with them. For example, the activation matrix generated by MUXQ can further incorporate the difficulty-migration strategy of SmoothQuant to achieve additional performance improvement.

The remainder of this paper is organized as follows. Section II describes the background of quantization and the role of outliers in quantization. Section III presents the MUXQ framework and explains how it reduces the influence of outliers. Section IV compares MUXQ, naive quantization, and LLM.int8() on GPT-2 models under different quantization settings. Finally, Section V concludes the paper and discusses the significance and future potential of MUXQ.

\begin{figure}[t]
    \centering
    \includegraphics[width=0.4\columnwidth]{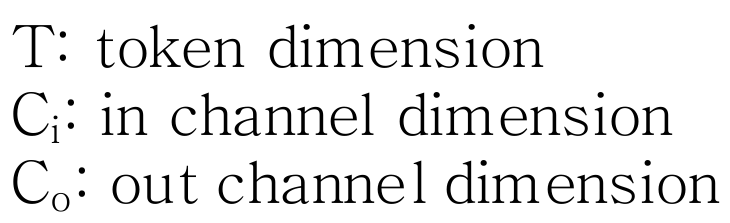}\\[0.7em]
    \includegraphics[width=1.0\columnwidth]{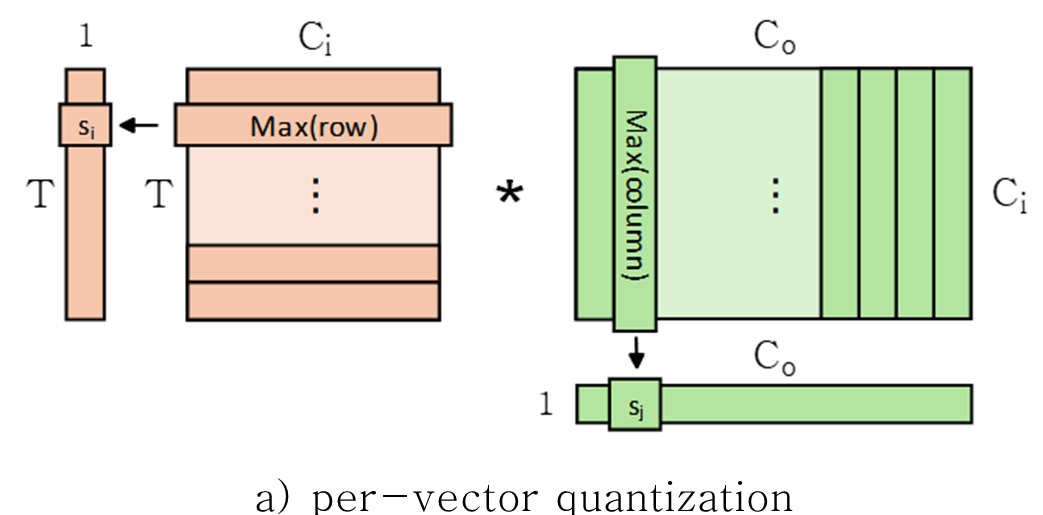}\\[0.7em]
    \includegraphics[width=1.0\columnwidth]{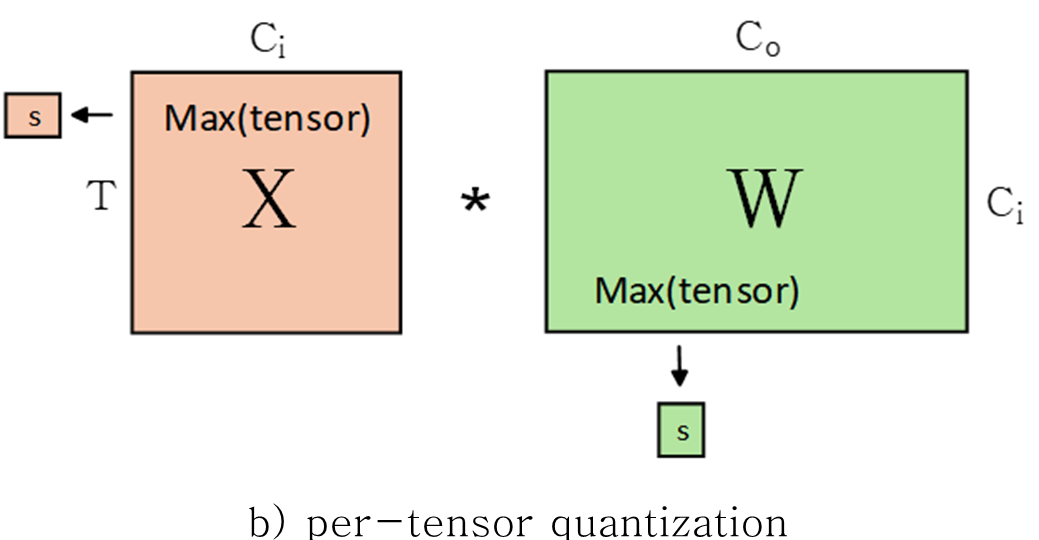}

    \caption{Quantization process for (a) per-vector quantization and (b) per-tensor quantization. In the per-vector case, activations and weights are quantized on a per-row or per-channel basis, respectively, and the scaling factor $s_i$ is determined by the maximum value of each corresponding vector.}
    \label{fig:three_vertical}
\end{figure}

\section{Background}

\subsection{Quantization}

Quantization refers to the process of mapping values represented in floating-point (FP) format to integer (INT) levels. Since INT operations are generally more efficient than FP operations on most hardware platforms, hardware efficiency can be improved by quantizing values, performing computation in INT, and then dequantizing the results afterward \cite{ref4}.

A basic quantization method is abs-max quantization, which quantizes values based on the element with the largest absolute magnitude within a given quantization unit. The following equations describe abs-max quantization. Specifically, the maximum absolute value within the quantization range is used to determine the scaling factor. The FP values are divided by this scaling factor and rounded to integers, and after integer matrix multiplication, the result is dequantized by multiplying the scaling factors back.

\begin{equation}
\begin{aligned}
\bar{X}_{\mathrm{INT8}}
&= \operatorname{round}\left(\frac{X_{\mathrm{FP16}}}{s_X}\right), \\
s_X
&= \frac{\max(|X_{\mathrm{FP16}}|)}{2^7 - 1}
\end{aligned}
\end{equation}

\begin{equation}
\begin{aligned}
\bar{W}_{\mathrm{INT8}}
&= \operatorname{round}\left(\frac{W_{\mathrm{FP16}}}{s_W}\right), \\
s_W
&= \frac{\max(|W_{\mathrm{FP16}}|)}{2^7 - 1}   
\end{aligned}
\end{equation}

\begin{equation}
\bar{Y}_{\mathrm{FP16}} = s_X \cdot s_W \cdot \left(\bar{X}_{\mathrm{INT8}} \cdot \bar{W}_{\mathrm{INT8}}\right)
\end{equation}

The quantization unit determines the granularity, depending on the range over which the scaling factor is shared. Granularity can be categorized into per-group, per-vector, and per-tensor quantization. In the per-vector setting, the quantization unit may correspond either to a token vector or a channel vector, which are referred to as per-token and per-channel quantization, respectively. Because per-group quantization incurs excessive overhead \cite{ref7}, the experiments in this paper consider only per-vector and per-tensor granularity. Figure 2 illustrates the quantization schemes under different granularity settings.

\subsection{Outlier handling}
Outliers have abnormally large magnitudes compared with other elements, which makes the scaling factor excessively small. Figure 3 illustrates how the presence of outliers prevents values from being distributed evenly. Therefore, if only the magnitudes of the outlier channels can be reduced to a level similar to that of normal channels, low-precision per-vector and per-tensor quantization can be enabled while preserving accuracy. Previous studies, as shown in Fig. 1, have observed that outlier features appear in a channel-wise manner. This suggests that quantizing activations on a per-channel basis could yield substantially higher accuracy \cite{ref8}.
\begin{figure}[h]
\centering
\includegraphics[width=1.0\linewidth]{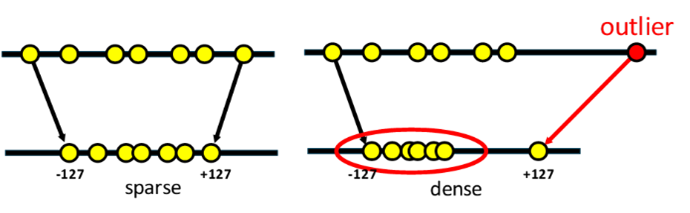}
\caption{\label{fig:outlier fig}Figure 3. The presence of outliers affects the scaling factor, causing the value distribution to become denser and thereby increasing quantization error.}
\end{figure}
However, per-channel activation quantization is difficult to support efficiently in GEMM kernels on hardware accelerators. For this reason, most prior works have adopted per-token or per-tensor activation quantization instead \cite{ref5}, \cite{ref6}. Taking this practical constraint into account, this work also adopts per-token and per-tensor granularity for activation quantization.

\begin{figure*}[h]
\centering
\includegraphics[width=1.0\linewidth]{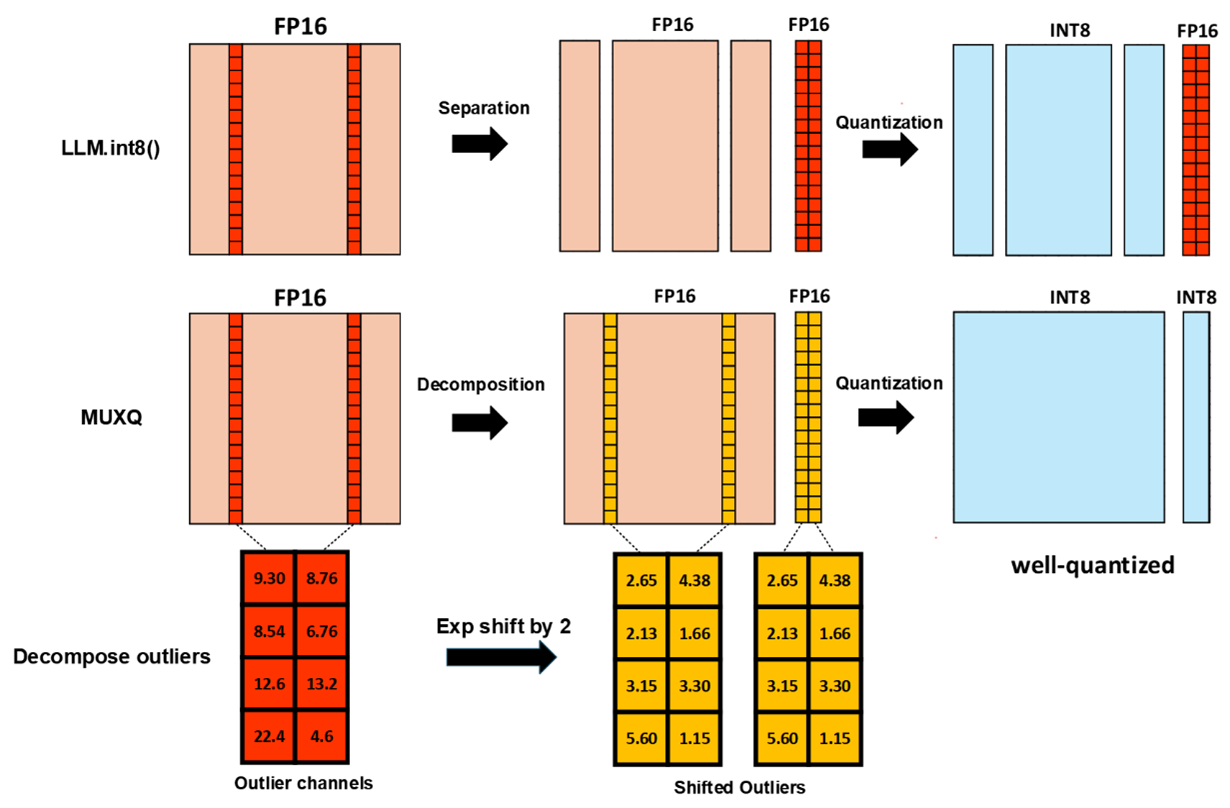}
\caption{\label{fig:muxq structure}Comparison between the MUXQ architecture and the LLM.int8() architecture. The lower part of the figure illustrates an example of outlier decomposition in MUXQ under the assumption that $\exp factor=2$.}
\end{figure*}

\section{MUXQ structure}

Given the channel-wise nature of activation outliers and their excessively large magnitudes, reducing the magnitudes of outlier channels is one promising way to address this problem. Based on this observation, we propose MUXQ (Mixed-to-Uniform Precision Matrix Quantization), which enables per-tensor quantization by redistributing the magnitudes of outlier channels through the addition of an auxiliary matrix to the activation matrix.

\subsection{Motivation}
In LLM fine-tuning, a widely used approach is to add a low-rank matrix to the weights and train only the added parameters \cite{ref9}. This method effectively reduces fine-tuning cost and remains popular in practice. Inspired by this idea, MUXQ was developed from the intuition that low-precision quantization may become feasible if a low-rank auxiliary matrix is added not to the weights but to the activations, thereby redistributing the magnitudes of activation outlier channels.
\subsection{Reducing Outlier Magnitudes}
A floating-point number consists of a sign, an exponent, and a mantissa. Among these components, the exponent has the greatest influence on magnitude. Therefore, if the exponent can be adjusted appropriately so that outlier channels have magnitudes comparable to those of normal channels, quantization error can be significantly reduced compared with naive quantization. However, applying such a correction to outlier channels would require an additional operation immediately after matrix multiplication, which is difficult to implement efficiently on hardware such as systolic arrays and MAC units. Rather than applying a post-multiplication correction to the outlier channels, hardware efficiency can be improved by adding a low-rank matrix and compensating for the outliers through a separately computed correction term.

\subsection{Outlier Decomposition method}
In MUXQ, after applying the proposed decomposition, we refer to the resulting main matrix and auxiliary low-rank matrix as \textit{Body} and \textit{Aux}, respectively. Here, $\mathrm{Body}_{\mathrm{outlier}}$ denotes the submatrix formed by the outlier columns of the Body matrix. Through this decomposition, the magnitude of each outlier column is no longer concentrated in a single activation column, but is instead distributed across the Body and Aux terms.

Let $X_{\mathrm{outlier}}$ denote the submatrix formed by the outlier columns of the input activation matrix $X$. MUXQ decomposes this outlier submatrix into two parts: a reduced main component and an auxiliary component. Specifically,
\begin{equation}
\mathrm{Body}_{\mathrm{outlier}} = X_{\mathrm{outlier}} \gg \exp
\end{equation}
\begin{equation}
\mathrm{Aux} = \mathrm{Body}_{\mathrm{outlier}}
\end{equation}
and the original outlier contribution is reconstructed as
\begin{equation}
X_{\mathrm{outlier}} = \mathrm{Body}_{\mathrm{outlier}} + (2^{\exp}-1)\mathrm{Aux}.
\end{equation}
In this way, MUXQ reduces the dynamic range of the main activation path while preserving the original numerical contribution of the outlier columns through the auxiliary path.

An important design issue in MUXQ is the choice of 
$\exp factor$, which determine how aggressively the outlier magnitude is redistributed. As noted earlier, MUXQ adopts the outlier criterion used in LLM.int8(), where an outlier channel is defined as a channel containing at least one element with magnitude greater than 6. Under this criterion, we set $\exp factor=2$ so that the outlier magnitude is reduced to a level comparable to that of normal channels.
However, this choice involves an implementation trade-off. As can be seen from the equations below,
\begin{equation}
\begin{aligned}
Y_{\mathrm{body}} &= \mathrm{Body} \cdot W, \\
Y_{\mathrm{Aux}} &= \mathrm{Aux} \cdot W, \\
Y_{\mathrm{total}} &= Y_{\mathrm{body}} + (2^{\exp} - 1)\,Y_{\mathrm{Aux}}
\end{aligned}
\end{equation}
when $\exp factor=1$, MUXQ can be implemented simply by summing the outputs of two matrix multiplications. In contrast, when $\exp factor$ are not equal to 1, additional overhead may be introduced. Therefore, $\exp factor$ should be selected by considering both quantization effectiveness and implementation cost.

\section{Experimental Setup}

\subsection{Evaluation method}
In this section, we construct an experimental setup to evaluate the effectiveness of the proposed MUXQ. For comparison, we evaluate naive quantization, MUXQ, and LLM.int8() under the same conditions. We progressively increase the difficulty of quantization and compare the resulting performance changes across methods. Model performance is evaluated using language modeling perplexity.

\subsection{Model and dataset}
We use the WikiText-2\cite{ref10} dataset for evaluation. WikiText-2 is a widely used benchmark for language modeling based on Wikipedia articles, making it suitable for evaluating language modeling performance across different models. The evaluation is conducted on the GPT-2 family, including GPT-2 small (0.1B), medium (0.3B), and large (0.7B). All models are pretrained checkpoints provided through Hugging Face and implemented using the Hugging Face Transformers framework.

\begin{table*}[t]
\centering
\caption{Perplexity comparison under different quantization settings.}
\label{tab:perplexity_comparison}
\resizebox{\textwidth}{!}{%
\begin{tabular}{lllcccccc}
\toprule
\multirow{2}{*}{model} & \multirow{2}{*}{granularity} & \multirow{2}{*}{target layers} & \multicolumn{2}{c}{bits} & \multicolumn{4}{c}{perplexity} \\
\cmidrule(lr){4-5} \cmidrule(lr){6-9}
& & & IA(bits) & W(bits) & naive & MUXQ & llm.int8() & fp16 \\
\midrule

\multirow{5}{*}{GPT2-small(0.1B)}
& \multirow{4}{*}{per-vector} &  & 8 & 8 & 25.3825 & 25.28   & 25.2493 & 25.188 \\
&                            &  & 7 & 8 & 25.8291 & 25.487  & 25.3867 & 25.188 \\
&                            &  & 6 & 8 & 30.2001 & 26.9466 & 26.1048 & 25.188 \\
&                            &  & 5 & 8 & 152.1251 & 43.129  & 30.5337 & 25.188 \\
\cline{2-2}\cline{4-9}
& per-tensor & attention & 8 & 8 & \textcolor{red}{50.618} & \textbf{29.5534} & 28.4595 & 25.188 \\
\cline{1-2}\cline{4-9}

\multirow{3}{*}{GPT2-medium(0.3B)}
& \multirow{3}{*}{per-tensor} & mlp & 8 & 8 & 22.5081  & 20.146  & 19.6749 & 18.4742 \\
&                              &     & 7 & 8 & 35.8196  & 22.3645 & 20.4703 & 18.4742 \\
&                              &     & 6 & 8 & \textcolor{red}{801.9368} & 38.2443 & 25.009  & 18.4742 \\
\cline{1-2}\cline{4-9}

\multirow{3}{*}{GPT2-large(0.7B)}
& \multirow{3}{*}{per-tensor} &  & 8 & 8 & 17.0662 & 16.7404 & 16.7041 & 16.4541 \\
&                             &  & 7 & 8 & 19.4031 & 17.2334 & 17.0956 & 16.4541 \\\cline{4-9}&                             &  & 6 & 8 & \textcolor{red}{40.0405} & \textbf{19.2885} & 18.7935 & 16.4541 \\

\bottomrule
\end{tabular}%
}
\end{table*}
\subsection{Implementation}
To minimize implementation complexity, we adopt the abs-max scaling method. Since scaling method is not treated as a primary experimental variable in this study, the same scaling method is used for all models. To incorporate the MUXQ structure, the original Conv1D modules in the Hugging Face framework are replaced with customized Conv1D modules. Quantization is applied to the main projection modules in GPT-2, including the attention modules ($c\_attn$ and the attention $c\_proj$) and the MLP modules ($c\_fc$ and the MLP $c\_proj$). However, instead of executing a true quantize-compute-dequantize inference pipeline, we evaluate accuracy using fake quantization\cite{ref11}, which follows a quantize-dequantize-compute procedure. Although fake quantization provides accuracy trends similar to those of simulated quantized inference, it does not reflect actual inference speed. Therefore, this study does not include an experimental comparison of inference latency. All experiments are conducted on an NVIDIA GeForce RTX 3080 (12 GB).
\subsection{Experimental Results}
Table 1 reports the perplexity results on the WikiText-2 dataset for the three GPT-2 models. Overall, MUXQ yields lower perplexity than naive quantization in most cases, although it generally remains slightly worse than LLM.int8(). The benefit of MUXQ in handling outliers is most clearly observed when activation precision is reduced. By contrast, reducing weight precision does not significantly affect the relative behavior of the outlier-handling methods, while varying quantization granularity enables a more detailed analysis and further confirms that MUXQ consistently outperforms naive quantization.

Among these factors, reducing activation precision plays the most important role in this study. Since the main purpose of the experiment is to evaluate the handling of activation outliers, the difference between MUXQ and naive quantization becomes more evident as activation precision decreases. As shown in Table 1, for all three GPT-2 models, once activations are quantized below a certain bit width, naive quantization suffers severe accuracy degradation, whereas MUXQ and LLM.int8() remain relatively stable. In particular, for GPT-2 medium, when activation precision is reduced to 7 bits, the perplexity of naive quantization rises sharply to above 35, while MUXQ and LLM.int8() maintain perplexity in the range of approximately 20 to 22. However, when activation precision is further reduced to 6 bits, the perplexity of MUXQ becomes substantially worse than that of LLM.int8(). This suggests that, as activation precision is pushed to lower bit widths, the FP16-based outlier handling of LLM.int8() becomes increasingly advantageous.

Table 2 shows that reducing weight precision results in performance changes of a similar magnitude across all three methods. This suggests that weight precision is not a major factor in distinguishing the three methods in this study, since both MUXQ and LLM.int8() primarily target activation outliers. In addition, as quantization granularity becomes finer, the baseline performance of the model improves significantly, indicating that the model becomes more robust to quantization. This makes it easier to clearly observe the advantage of MUXQ when activation precision is reduced. For example, in the GPT-2 large results with 6-bit activation precision, MUXQ maintains performance comparable to that of LLM.int8(), whereas naive quantization exhibits severe degradation, with perplexity increasing to more than twice its original value.

\begin{table}[ht]
\centering
\caption{Perplexity comparison under different bit settings.}
\label{tab:small_perplexity}
\resizebox{\columnwidth}{!}{%
\begin{tabular}{cccccc}
\toprule
\multicolumn{2}{c}{} & \multicolumn{4}{c}{perplexity} \\
\cmidrule(lr){3-6}
IA(bits) & W(bits) & naive & MUXQ & llm.int8() & fp16 \\
\midrule
8 & 5 & 27.5149 & 27.3616 & 26.5359 & 25.188 \\
8 & 4 & 41.6420 & 41.1245 & 35.4289 & 25.188 \\
\bottomrule
\end{tabular}%
}
\end{table}

\subsection{Discussion}

The goal of quantized model inference is to achieve faster inference through quantization without significantly sacrificing accuracy. In this study, we show that MUXQ can preserve accuracy relative to other quantization methods, but we do not experimentally evaluate inference speed. Nevertheless, because INT8 matrix multiplication is generally more efficient than FP16 matrix multiplication, MUXQ may offer practical deployment advantages in INT-oriented hardware environments, although this should be validated through kernel-level latency and power measurements on real hardware.
\section{Conclusion}
In this work, we proposed MUXQ to alleviate the activation-outlier problem in low-precision quantization of large language models. While existing methods either rely on mixed-precision computation at the cost of hardware efficiency to preserve accuracy or fail to sufficiently address outlier channels, MUXQ redistributes outlier magnitudes within INT precision to produce a more balanced activation distribution. In doing so, it achieves a better balance between accuracy and hardware efficiency.

MUXQ introduces an auxiliary matrix into the activation matrix to mitigate the magnitudes of channels in which outliers are concentrated, thereby achieving better performance than naive quantization. In addition, because the MUXQ structure does not conflict with other quantization approaches, it can be readily combined with additional techniques such as SmoothQuant or other calibration-based methods.

In summary, although MUXQ introduces a small amount of additional computation, it maintains low overhead and provides stable LLM inference performance even under uniform INT quantization. Future work may extend MUXQ by combining it with other quantization methods or by validating its speed and power efficiency on real hardware accelerators, ultimately moving toward practical on-device INT-only LLM acceleration.

\end{document}